%% file: colm2025_conference.tex
\documentclass{article} % For LaTeX2e
\usepackage[preprint]{colm2025_conference}

\usepackage{microtype}
\usepackage{hyperref}
\usepackage{url}
\usepackage{booktabs}

\usepackage{lineno}

\usepackage[utf8]{inputenc} % allow utf-8 input
\usepackage[T1]{fontenc}    % use 8-bit T1 fonts

\usepackage{amsfonts}       % blackboard math symbols
\usepackage{amsmath}
\usepackage{nicefrac}       % compact symbols for 1/2, etc.
\usepackage{microtype}      % microtypography
\usepackage{xcolor}         % colors

\usepackage{array}
\usepackage{colortbl}
\usepackage{wrapfig}
\usepackage{graphicx}
\usepackage{cleveref}
\usepackage{enumitem,kantlipsum}
\usepackage{tikz}
\usetikzlibrary{arrows.meta, positioning, shapes.geometric, shadows}

\usepackage{tcolorbox}
\tcbuselibrary{listingsutf8}

\definecolor{darkblue}{rgb}{0, 0, 0.5}
\hypersetup{colorlinks=true, citecolor=darkblue, linkcolor=darkblue, urlcolor=darkblue}

\usepackage{subcaption}

\newcommand{\ZSEVAL}{\textsc{ZeroSumEval}}

\definecolor{zse_color1}{RGB}{41,44,147}
\definecolor{zse_color2}{RGB}{14,140,247}

\title{\ZSEVAL: Scaling LLM Evaluation with Inter-Model Competition}

\author{Haidar Khan\thanks{Core contributors. Correspondence to \texttt{haidark@meta.com}.} \\
Meta \\
\And
Hisham A. Alyahya\footnotemark[1] \\
Saudi Data \& AI Authority \\
\And
Yazeed Alnumay \\
Cohere \\
\And
M Saiful Bari \\
Saudi Data \& AI Authority \\
\And
Bülent Yener \\
Rensselaer Polytechnic Institute \\
}

\begin{document}

\ifcolmsubmission
\linenumbers
\fi

\maketitle

\begin{abstract}
\input{sections/00_abstract}
\end{abstract}

\section{Introduction}
\input{sections/10_introduction}

\section{Methodology}
\input{sections/30_methodology}

% \section{Implementation}
% \input{sections/40_implementation}

\section{Experiments}
\input{sections/50_experiments}

\section{Results}
\input{sections/60_results}

\section{Related Work}
\input{sections/20_related_work}

\section{Conclusion}
\input{sections/70_conclusion}

\subsubsection*{Acknowledgments}

The authors thank Colton Ritchie for contributions to security game aspects of \ZSEVAL{} and early versions of this work.

\bibliography{colm2025_conference}
\bibliographystyle{colm2025_conference}

\appendix
\section{Appendix}

\input{sections/90_appendix}

\end{document}

%% file: sections/00_abstract.tex
Evaluating the capabilities of Large Language Models (LLMs) has traditionally relied on static benchmark datasets, human assessments, or model-based evaluations - methods that often suffer from overfitting, high costs, and biases. ZeroSumEval is a novel competition-based evaluation protocol that leverages zero-sum games to assess LLMs with dynamic benchmarks that resist saturation. ZeroSumEval encompasses a diverse suite of games, including security challenges (PyJail), classic games (Chess, Liar's Dice, Poker), knowledge tests (MathQuiz), and persuasion challenges (Gandalf, Debate). These games are designed to evaluate a range of AI capabilities such as strategic reasoning, planning, knowledge application, and creativity. Building upon recent studies that highlight the effectiveness of game-based evaluations for LLMs, ZeroSumEval enhances these approaches by providing a standardized and extensible framework. To demonstrate this, we conduct extensive experiments with >7000 simulations across 7 games and 13 models. Our results show that while frontier models from the GPT and Claude families can play common games and answer questions, they struggle to play games that require creating novel and challenging questions. We also observe that models cannot reliably jailbreak each other and fail generally at tasks requiring creativity. We release our code at \url{https://github.com/facebookresearch/ZeroSumEval}.

%% file: sections/10_introduction.tex
\label{sec:introduction}

% - LLMs are being rapidly developed
% - More and more often, expensive choices need to get made on which model to use
% - Benchmarking is the way to make these choices
% - Current benchmarks have issues (contamination, sensitivity, non-diversity, high cost of creating new benchmarks, benchmarks often becoming obsolete)
% - Prior work proposed games as benchmarks, which seems promising
% - These benchmarks are inflexible
% 	- Prone to prompt sensitivity
% 	- Too small of a set of games
% 	- Difficult to extend with more games
% 	- Doesn’t allow for different prompting strategies
% 	- Difficult to interpret
% 	- No easy way to give feedback to the model when making illegal moves
% - \ZSEVAL{} attempts to solve these issues
% - Contributions:
% 	- Flexible and extensible open source framework
% 	- Robust to prompt sensitivity
% 	- Easy to interpret the behavior of models
% 	- Models are given multiple chances to make a legal move

% \textcolor{red}{A paragraph about LLM advancements and the need for meaningful evaluations.}

Large Language Models (LLMs) are being developed at an unprecedented pace \citep{zhao2024surveylargelanguagemodels}, requiring significant investment for their training and refinement \citep{metagenai2024,techcrunch2022,cnbc2024}. As the performance and complexity of these models continue to grow \citep{chen2024chatgptsoneyearanniversaryopensource}, selecting the most appropriate model for a specific application has become an increasingly challenging and costly decision~\citep{kaplan2020scalinglawsneurallanguage,hoffmann2022trainingcomputeoptimallargelanguage}. Benchmarking emerges as critical in this context \citep{laskar2023systematic,qin2023chatgpt}, providing standardized metrics and evaluations to guide these choices.

% With the rapid growth of generative technologies built on top of Large Language Models \citep{chatgptblog,geminiteam2024gemini,anthropic2024claude,rekateam2024reka,mistralnews,llama3,qwen2}, it has been increasingly difficult to evaluate these models comprehensively \citep{guo2023evaluatinglargelanguagemodels}. 
Current benchmarking practices face several significant issues. Many benchmarks suffer from data contamination \citep{yang2023rethinking}, where models inadvertently train on portions of the test data \citep{llama3,groeneveld2024olmoacceleratingsciencelanguage}, leading to inflated performance metrics. Sensitivity to prompt variations \citep{alzahrani2024benchmarks} and a lack of diversity in evaluation tasks \citep{laskar2024systematic} further undermine the reliability and robustness of these benchmarks. Additionally, the high cost and effort required to develop new benchmarks often result in outdated evaluation methods that do not keep pace with the rapid development of LLMs \citep{kiela2021dynabenchrethinkingbenchmarkingnlp,vu2023freshllms}.

\begin{figure}[ht!]
  \begin{center}
    \includegraphics[width=\linewidth]{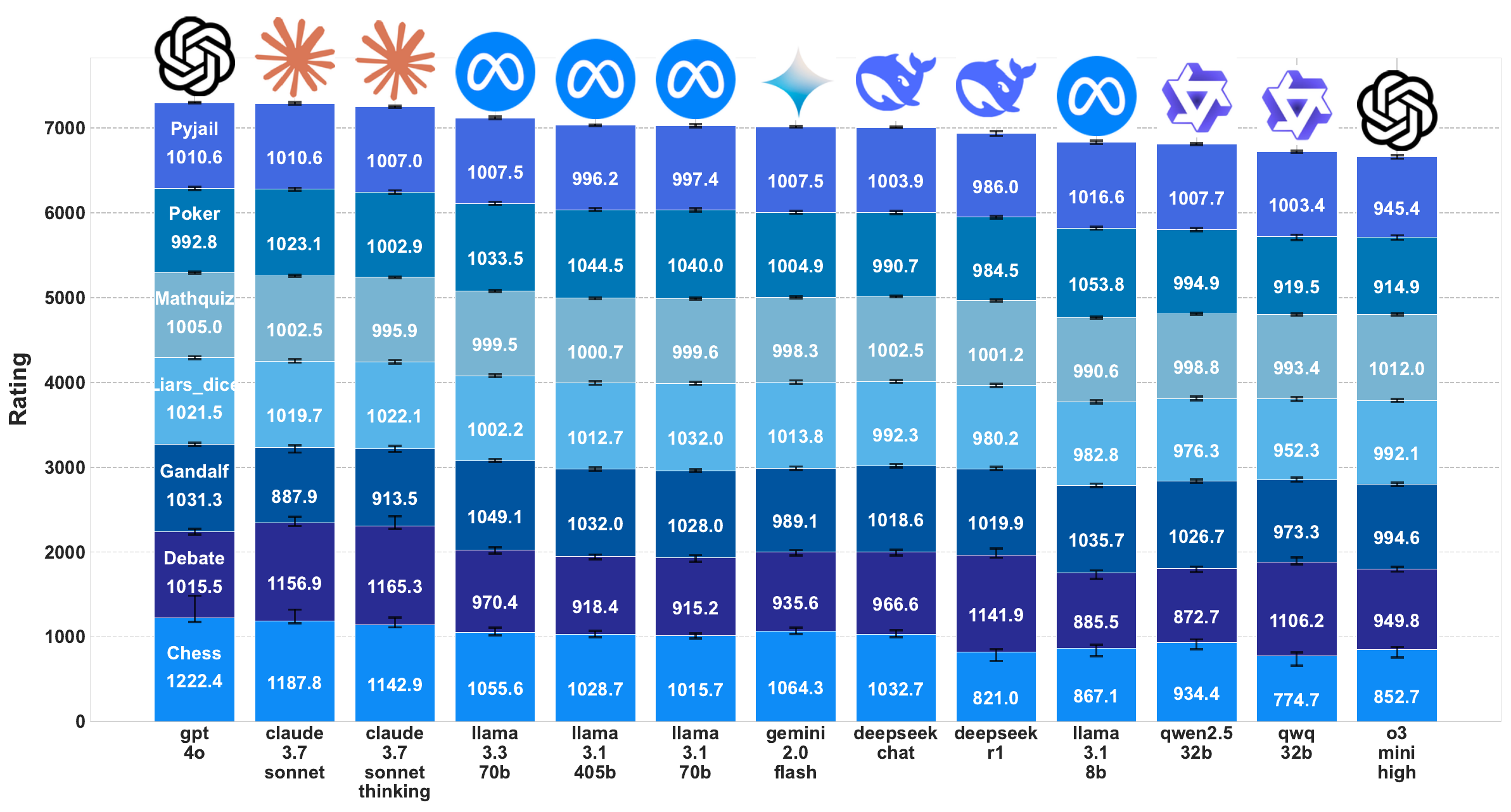}
  \end{center}
  \caption{Cumulative ratings of 13 models on \ZSEVAL{}. The top performing models (\texttt{gpt-4o} and \texttt{claude-3.7-sonnet}) show mostly on-par performance across \ZSEVAL{} games. Thinking model quality varies between model families (e.g. \texttt{claude-3.7-sonnet-thinking} vs \texttt{deepseek-r1}). Surprisingly, \texttt{o3-mini-high} performs worst amongst this cohort of models.}
  \label{fig:cumulative_ratings}
\end{figure}

An observed disparity exists between the computational resources measured in floating-point operations per second, or FLOPs used to train LLMs and those allocated for their evaluation. Training these models involves massive computational efforts \citep{hoffmann2022trainingcomputeoptimallargelanguage}, yet the evaluation phase typically utilizes a negligible fraction of this capacity \citep{laskar2024systematic}. Scaling up evaluation by increasing the number of evaluation tokens is essential for a more thorough understanding of model capabilities. Traditionally, this scaling involves incorporating human-crafted data \citep{holland2018datasetnutritionlabelframework}, which is resource-intensive \citep{hutchinson2021accountabilitymachinelearningdatasets} and may not adequately capture the complexities of language \citep{10.1145/3457607} and reasoning required to challenge advanced LLMs \citep{gudibande2023falsepromiseimitatingproprietary}.

Previous work has proposed the use of games as benchmarks \citep{topsakal2024evaluatinglargelanguagemodels}, offering a promising avenue for evaluating complex reasoning \citep{wong2023wordmodelsworldmodels} and decision-making abilities of LLMs \citep{conll-2023-babylm,park2023generative,wang2023voyageropenendedembodiedagent}. Games provide interactive and dynamic environments that can test models beyond static datasets. However, existing game-based benchmarks are often \emph{(i)} inflexible and limited in scope, \emph{(ii)} not easily extendable, \emph{(iii)} restricted in their effectiveness for comprehensive model evaluation, and \emph{(iv)} depend on predefined prompts.

Scaling evaluation is fundamental not only for assessing performance but also for uncovering hidden dynamics within LLMs, such as potential backdoors or biases \citep{schuster-etal-2020-limitations}, and for evaluating their emerging reasoning capabilities \citep{brown2020languagemodelsfewshotlearners,sanh2022multitaskpromptedtrainingenables,wei2023chainofthoughtpromptingelicitsreasoning,wei2023inversescalingushaped}. Implementing environments for simulations or games offers a scalable solution to these challenges \citep{openai2019solvingrubikscuberobot,OpenAI2019,silver2016mastering,silver2017masteringchessshogiselfplay,zheng2021aieconomistoptimaleconomic}.

Existing evaluation protocols possess several key issues:

\begin{itemize}[wide, labelwidth=!, labelindent=0pt, itemindent=!]

\item \textbf{Limited Diversity}: Traditional evaluation methods rely on static datasets, which are inherently limited by their dependency on human curation and annotation. This makes it challenging to continuously introduce new, diverse test data. 
% An extensible simulated environment, however, allows for a wide array of dynamically generated games and scenarios, enhancing the diversity and scalability of evaluation tasks.

\item \textbf{Crowd and Annotator Bias}: LLM evaluations conducted by large crowds often tend to be susceptible to social hacking, and it can depend on geographic, temporal, and narrative factors \citep{gururangan-etal-2018-annotation}. Controlled and interpretable environments can mitigate these biases by providing consistent, objective evaluation criteria.

\item \textbf{Sensitivity to Prompts}: Previous work \citep{zheng2024largelanguagemodelsrobust,pezeshkpour2023largelanguagemodelssensitivity,lu-etal-2022-fantastically,alzahrani2024benchmarks,wang2024answersreviewingrationalitymultiple} has shown that models are sensitive to benchmark formats. 
% By sheer chance, a model could be presented with a prompting method that's either favorable or detrimental. 
These prompt modifications are shown to result in substantially different relative performance between models \citep{alzahrani2024benchmarks}. 
% By testing models in varied scenarios within a controlled environment, we can assess and improve their robustness to different prompts. Crucially, different models are not optimized for the same prompts due to variations in data mixtures and algorithmic implementations. Using identical prompts across all models may therefore lead to unfair comparisons.

% \emph{(iv)} \textbf{Interpretability}: Controlled environments significantly enhance our ability to interpret model behaviors and decision-making processes. By providing a structured, programmable framework for evaluation, these environments allow for detailed analysis of model outputs across various scenarios. 
% % This programmatic approach facilitates a tight feedback loop, enabling model developers to swiftly iterate on their designs. 
% The ability to rapidly test hypotheses, identify shortcomings, and implement improvements accelerates the overall development process. Moreover, the systematic nature of these evaluations provides valuable insights into the model's underlying mechanisms, potentially revealing emergent behaviors or unexpected capabilities.

\item \textbf{Saturation}: With the rapid improvement of LLMs, evaluation benchmarks quickly become obsolete and saturated, with frontier models achieving almost perfect scores, which necessitates the development of new benchmarks. 
% On the opposite extreme, benchmarks that are too difficult would result in almost random scores. Both extremes result in a lack of granularity to distinguish models. Therefore, benchmarks posing moderate difficulty to frontier models will need to be continuously developed as models improve. 
For instance, GSM8K \citep{gsm8k} tests models on grade school-level math, and most state-of-the-art models achieve scores above 90\% \citep{llama3, claude_35_sonnet}. Thus, the more difficult MATH \citep{hendrycksmath} dataset, which consists of math competition questions, was developed and is now commonly used \footnote{\label{hf_leaderboard}HuggingFace's \href{https://huggingface.co/spaces/open-llm-leaderboard/open_llm_leaderboard}{Open LLM Leaderboard} \citep{open-llm-leaderboard-v1, open-llm-leaderboard-v2} migrated from GSM8K in v1 to MATH in v2.}. A similar trend is observed in academic examination benchmarks with the migration from MMLU \citep{mmlu} to MMLU-Pro \citep{mmlu_pro} and GPQA. \citep{gpqa}\footnote{Similar to \ref{hf_leaderboard}, the leaderboard transitioned from MMLU in v1 to MMLU-Pro and GPQA in v2.}. 
% Our interactive zero-sum game structure naturally scales without intervention, as models compete against other better models.

\end{itemize}

To address these challenges, we introduce \ZSEVAL{}, an extensible open-source framework designed to evaluate LLMs through the simulation of multi-player zero-sum games. The evaluations in the framework naturally scale and resist saturation as model capabilities improve. 
% \ZSEVAL{} adopts a programming level abstraction of prompting~\citep{khattab2023dspy} to reduce prompt sensitivity. This allows for robust assessment by providing models with multiple opportunities to make legal moves and accommodates occasional errors to further reduce the effect of prompt formatting.

In this work, we show that \ZSEVAL{} possesses the following properties:

\begin{itemize}[wide, labelwidth=!, labelindent=0pt, itemindent=!]

\item \textbf{Scalable Evaluation}: We demonstrate how simulation environments can effectively scale the evaluation process.

\item \textbf{Dynamic Evaluation}: Each game trace explores a different path in an almost infinite space of move trajectories. Regurgitation of memorized information is penalized naturally by threat of exploitation from opposing players.

\item \textbf{Robustness to Prompt Sensitivity}: Prompt writing is abstracted to the input/output level for all models and feedback loops are used to eliminate the need for elaborate and detailed prompts.

\item \textbf{Enhanced Interpretability}: Game traces reveals the strengths and weaknesses of models in ways static benchmarks cannot.

\end{itemize}

%% file: sections/30_methodology.tex
In this section, we describe the technical details of \ZSEVAL{} including design choices, the game selection (including novel game types), and how prompting is abstracted. At its core, \ZSEVAL{} provides controlled environments to observe models competing against each other to win competitive games. In particular, ZSE controls (i) the role and information each model has access to at any point in the simulation and (ii) the feedback models can use to correct mistakes and act successfully. 

% \subsection{Capabilities}

% The games within ZSE are designed to evaluate specific capabilities in a controlled environment:

% \begin{itemize}%[wide, labelwidth=!, labelindent=0pt, itemindent=!]

% \item \textbf{Reasoning}: Board games and cybersecurity scenarios require models to perform complex, multi-step reasoning. They test the models' ability to process information, predict outcomes, and formulate strategies in dynamically changing environments.

% \item \textbf{Planning}: Board games also involve long-term strategy, requiring models to anticipate the consequences of their actions several moves ahead. This assesses the model’s foresight, adaptability, and capacity for nuanced decision-making.

% \item \textbf{Knowledge Application}: Models must recall and apply mathematical knowledge to solve problems in question answering type games. This setup provides a direct assessment of the models' ability to retrieve, interpret, and implement factual information in structured problem-solving.

% \item \textbf{Creativity}: Models successful at most types games must exhibit creativity to overcome opponents, especially in open-ended games like debating and jailbreaking.

% \end{itemize}

\subsection{Game Design}

\ZSEVAL{} supports an expanding suite of game types designed to test any aspect of LLM capabilities. The mix we showcase includes both well-known and established games, such as Chess, Poker, and Liar's Dice, as well as more special-purpose games (e.g. MathQuiz, Pyjail, and Gandalf). 
The following set of games are selected to encompass a range of capabilities, including strategic reasoning, planning, knowledge application, and creativity:

\begin{figure}[ht]
  \begin{center}
    \begin{tcolorbox}[title=Chess Game Trace, colback=gray!5, colframe=black!80, fonttitle=\bfseries]
\textbf{Position (FEN):} \\
\texttt{r1b1k2r/2pp1ppp/p2N1n2/npB1q3/8/2N5/PPPQPPPP/R3KB1R b KQkq - 8 11}

\medskip
\textbf{Move History:} \\
Nf3, e5, d4, exd4, Nxd4, Nf6, Nc3, Bc5, Nb3, a6, Be3, b5, Bxc5, Qe7, Nd4, Nc6, Nf5, Qe5, Qd2, Na5, \textbf{Nd6+}

\medskip
\textbf{Last Move:} MakeMove \textbf{(Nd6+)} \\
\textbf{White Rationale:}
\begin{itemize}
   \item We need to address the threat on the knight at f5 and improve our piece coordination.
  \item Moving the knight to \textbf{d6} would be a strong move, as it forks the black king and rook.
  \item This move creates immediate tactical pressure on black.
  \item It also activates white's pieces and opens up the position.
\end{itemize}
\medskip
\textbf{Player to Move:} Black (\texttt{llama3.3-70b}) \\
\textbf{Next Action:} MakeMove
    \end{tcolorbox}
  \end{center}
  \caption{Example of a chess game trace. Here \texttt{deepseek-chat} (white) executes a knight fork against \texttt{llama3.3-70b} (black) after which black loses the game by failing to produce a legal move.}
  \label{fig:chess_example}
\end{figure}

\paragraph{Classic Games (Chess, Poker, Liar's Dice)} Classic board games like chess serve as a benchmark for strategic reasoning and long-term planning. They require models to engage in multi-step thinking, manage trade-offs, and foresee opponent moves. This category is useful in evaluating a model's ability to plan several moves ahead, adapt its strategies, and make complex decisions under uncertainty\footnote{Chess has a rich history as a testbed for strategy and planning. See \url{https://github.com/carlini/chess-llm} and \url{https://huggingface.co/spaces/mlabonne/chessllm} for examples of LLMs playing chess.}. A snippet of a Chess game trace is shown in \Cref{fig:chess_example}.

\begin{figure}[ht]
  \begin{center}
    \begin{tcolorbox}[title=MathQuiz Game Trace, colback=gray!5, colframe=black!80, fonttitle=\bfseries]

\textbf{Target Answer:} 439

\medskip

\textbf{Teacher Question:} Find the smallest prime number greater than 400 that leaves a remainder of 1 when divided by 6 and a remainder of 5 when divided by 7.

\medskip
\textbf{Last Move:} TeacherAnswer \textbf{(439)} \\
\textbf{Teacher Rationale:}
\begin{itemize}
  \item Numbers 1 (mod 6) are of the form $6k + 1$: 1, 7, 13, 19, 25, ...
  \item Numbers 5 (mod 7) are of the form $7m + 5$: 5, 12, 19, 26, ...
  \item Find the smallest $n > 400$ such that:
    \begin{align*}
      n &\equiv 1 \pmod{6} \\
      n &\equiv 5 \pmod{7}
    \end{align*}
  \item First such value: $n = 19$, then add $\text{lcm}(6,7) = 42$ repeatedly:
  
    $19, 61, 103, \dots, 397, 439$
  \item Check if 439 is prime: not divisible by primes $\leq \sqrt{439} \approx 21$
  \item Therefore, \textbf{439 is prime} and satisfies both conditions.
\end{itemize}
\medskip
\textbf{Next Action:} StudentAnswer
    \end{tcolorbox}
  \end{center}
  \caption{Example of a MathQuiz game trace. At this point in the game, \texttt{claude-3.7-sonnet} (teacher) has generated a question and proven the question is valid by solving it. Now it is \texttt{llama3.1-405b}'s (student) turn to answer the question (which it fails to do after multiple attempts).}
  \label{fig:mathquiz_example}
\end{figure}

\paragraph{Question-Answer Games (MathQuiz)} These games are constructed to measure models' knowledge recall and logical reasoning abilities. MathQuiz, for instance, challenges models to both create and answer arithmetic and mathematical questions, assessing their understanding of mathematical concepts, computational accuracy, and step-by-step problem-solving skills (see \Cref{fig:mathquiz_example} for an example). Our implementation of MathQuiz tasks a teacher player to create a challenging math problem and prove that the problem is valid and solvable. A student player then attempts to answer the generated math problem. The student wins the game by answering the question correctly or if the teacher fails to create a valid question.

\paragraph{Conversational Games (Debate, Gandalf)} These are games designed to assess models' creative capabilities. For example, Gandalf engages LLMs in scenarios where one player attempts to break the conditioning/instruction following of another. In this game, a sentinel player is tasked to keep a randomly generated passphrase secret while conversing with an infiltrator player. The infiltrator wins if the passphrase is revealed in the conversation and the sentinel wins if the conversation exceeds a maximum number of turns~\citep{lakera2025gandalf}.

\paragraph{Cybersecurity Games (PyJail)} PyJail involves python ``capture the flag'' cybersecurity challenges, targeting the model's ability to create puzzles and interact with a restricted python environment to strategize solutions. Similar to MathQuiz, PyJail is structured into three stages: generating a challenge, proving the challenge validity, and posing the challenge to the opponent.

Details about these games can be found in \Cref{game_details}.

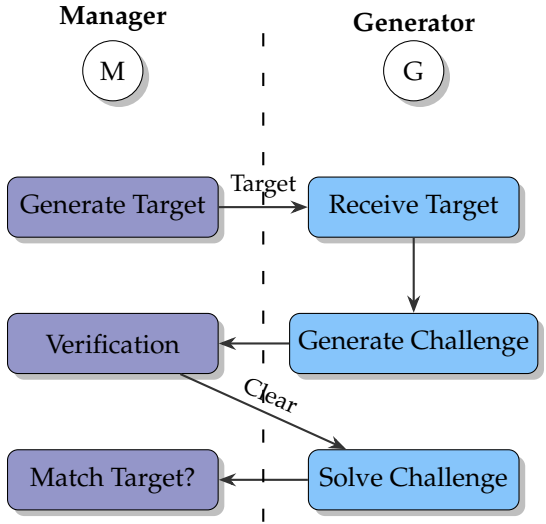
\begin{wrapfigure}[25]{R}{0.5\linewidth}
\vspace{-1em}
    \begin{center}
    \begin{tikzpicture}[
        box/.style={rectangle, draw, rounded corners, fill=zse_color1!50, drop shadow, minimum height=0.8cm, minimum width=2.8cm, align=center},
        boxllm/.style={rectangle, draw, rounded corners, fill=zse_color2!50, drop shadow, minimum height=0.8cm, minimum width=2.8cm, align=center},
        actor/.style={circle, fill=white!10, draw, drop shadow, minimum size=0.8cm},
        arrow/.style={-Stealth, thick, draw=black!80},
        node distance=1cm and 2cm
    ]
    % Actors
    \node[actor, label=above:{\textbf{Manager}}] (manager) at (-2, 0.5) {M};
    \node[actor, label=above:{\textbf{Generator}}] (generator) at (2, 0.5) {G};
    
    % Nodes for Manager
    \node[box, below=of manager] (generate) {Generate Target};
    \node[box, below=of generate] (verify) {Verification};
    \node[box, below=of verify] (match) {Match Target?};
    
    % Nodes for Generator
    \node[boxllm, below=of generator] (genModel) {Receive Target};
    \node[boxllm, below=of genModel] (output) {Generate Challenge};
    \node[boxllm, below=of output] (clear) {Solve Challenge};
    
    % Arrows between Manager and Generator
    \draw[arrow] (generate) -- (genModel) node[midway, above, font=\small] {Target};
    \draw[arrow] (genModel) -- (output);
    \draw[arrow] (output) -- (verify);
    \draw[arrow] (verify) -- (clear) node[midway, above, sloped, font=\small] {Clear};
    \draw[arrow] (clear) -- (match);
    
    % Dashed line separating roles
    \draw[dashed, thick, dash pattern=on 5pt off 10pt] (0, 1) -- (0, -5.5);
    \end{tikzpicture}
    \end{center}
    \caption{State diagram of the verification process involving the ZSEval Manager and the LLM Generator. Blue boxes indicate deterministic steps and green boxes indicate steps involving the LLM.}
    \label{fig:state-diagram-verification}
\end{wrapfigure}

\subsection{Scalable Verification}
\label{sec:scalable_verification}
The MathQuiz and PyJail games require competing models to generate \textbf{valid} and \textbf{solvable} challenges. Since verification of the challenges by a human in the loop is not scalable, we design a method to verify model output using an automated manager in a two-fold generation and verification process. This is accomplished by defining a target outcome (e.g., the answer to a math question or a CTF flag) as the basis for verifying the generated challenge, and regulating the model context to prevent target leakage.

The exact process (shown in \Cref{fig:state-diagram-verification}) is best illustrated with an example from MathQuiz (\Cref{fig:mathquiz_example}):

1. \ZSEVAL{} Manager generates a random integer from \texttt{(0,1000]}  as the target answer (439) and instructs the teacher (\texttt{claude-3.7-sonnet}) to generate a question.

2. The teacher \texttt{claude-3.7-sonnet} generates the question "Find the smallest prime number greater than 400 that leaves a remainder of 1 when divided by 6 and a remainder of 5 when divided by 7."

3. Ensuring the target answer is not in the context, \ZSEVAL{} Manager asks \texttt{claude-3.7-sonnet} to answer its own question and verifies against the target answer. If this step succeeds, the game continues. Without access to the answer, there is a $1/1000$ chance the teacher will produce the target answer by randomly guessing.

4. \ZSEVAL{} Manager asks the student (\texttt{llama-3.1-405b}) to answer the question.

This method ensures the generated challenge is valid and a solution is proven possible. The design also correctly penalizes models that directly generate memorized questions as they are likely to have been memorized by other models, thereby models that create challenging and novel questions are more likely to win. 

\subsection{Prompt Abstraction and Feedback}

We leverage the DSPy \citep{khattab2023dspy} approach to prompt abstraction in our framework. 
% DSPy provides a level of abstraction that we argue to be valuable for the implementation of move-making strategies. 
% We find its approach of abstracting away prompts to focus on the underlying logical structure of the pipeline to be ideal for 
DSPy provides abstractions that allow easily swapping and modifying strategies without depending on manually-written prompt templating. Another important reason for using DSPy is to make use of DSPy Assertions \citep{singhvi2024dspyassertionscomputationalconstraints}. This functionality simulates interactivity between the models and the game environment by allowing a number of retries (with feedback) when the model makes an invalid move. 
% Although we find that DSPy has the flexibility and generalizability to support various models and games, 
\ZSEVAL{} also supports arbitrary functions as player strategies, which allows for the use of frameworks like LangChain~\citep{langchain} and others.
% or the implementation of move making that is not based on LLMs (for example, a chess engine like Stockfish or human input from the console).

\subsection{Ratings}
\label{ratings}
Following recent suggestions for LLM rating systems by \citet{boubdir2023elouncoveredrobustnessbest,eloarena2023}, \ZSEVAL{} employs the Bradley-Terry (BT) rating system, an alternative to the Elo system, to rate models. The BT model is permutation-invariant and assumes a fixed win rate for each model pair, maximizing the likelihood of observed outcomes \citep{bradleyterryrank}. This choice is more suitable than the traditional Elo system, which was designed for human chess players with changing skill levels, whereas LLMs have fixed skill levels defined by their weights \citep{elo}.

%% file: sections/50_experiments.tex
% - Stability: 5 runs without optimization
% - Prompt Sensitivity: 5 runs with optimization
% - Absolute vs Relative: Compare to SOTA benchmark rankings

% Experimental setup

% - models tested

% - game settings

% - DSPy modules and optimizers

% - Matchings (round robin)

% - compiling tournament data

In this section, we describe the experiments to demonstrate the effectiveness of the \ZSEVAL{} as a benchmark. We also design experiments to compare prompting strategy (Predict vs Chain of Thought (CoT)) and compare "thinking" models against their instruct counterparts prompted with CoT. 

\subsection{Model Selection and Experimental Setup}
\begin{table}[h]
  \centering
  \begin{tabular}{l|l|l}
    \toprule
    \textbf{Family} & \textbf{Instruct Models} & \textbf{Thinking Models} \\
    \midrule
    OpenAI & \texttt{gpt-4o-2024-11-20} & \texttt{o3-mini-high-2025-01-31} \\
    Anthropic & \texttt{claude3.7-sonnet-20250219} & \texttt{claude3.7-sonnet-20250219-thinking} \\
    Meta & \texttt{llama3.1-8b}, \texttt{llama3.1-70b},  & -- \\
     & \texttt{llama3.1-405b}, \texttt{llama3.3-70b}, & \\
     & all \texttt{instruct} & \\
    Deepseek & \texttt{deepseek-v3-12-24} & \texttt{deepseek-r1-01-25} \\
    Google & \texttt{gemini-2.0-flash-001} & -- \\
    Qwen & \texttt{qwen2.5-32b-instruct} & \texttt{qwq-32b} \\
    \bottomrule
  \end{tabular}
  \caption{Models included in our initial rankings. Accessed via \url{https://openrouter.ai/}.}
  \label{tab:models_summary}
\end{table}

We select thirteen models of varying sizes and capabilities from six model families for this study (detailed in \Cref{tab:models_summary}).

In each experiment we conduct multiple round-robin tournaments to simulate competitive gameplay among the model (500-1500 games per experiment). 
Each tournament round involves all possible match permutations between model variants, after which the models' ratings are calculated (as described in \Cref{ratings}).

We set $\texttt{max\_player\_attempts}=3$ for Debate, Gandalf, and Liar's Dice as these games are easy for models to play. For Chess, MathQuiz, PyJail, and Poker we set $\texttt{max\_player\_attempts}=5$. 

%% file: sections/60_results.tex
We begin by noting a few general observations from the experiments and follow with specific learnings from the overall ratings and model performance comparisons.

\begin{figure}[ht]
  \begin{center}
    \includegraphics[width=\linewidth]
    {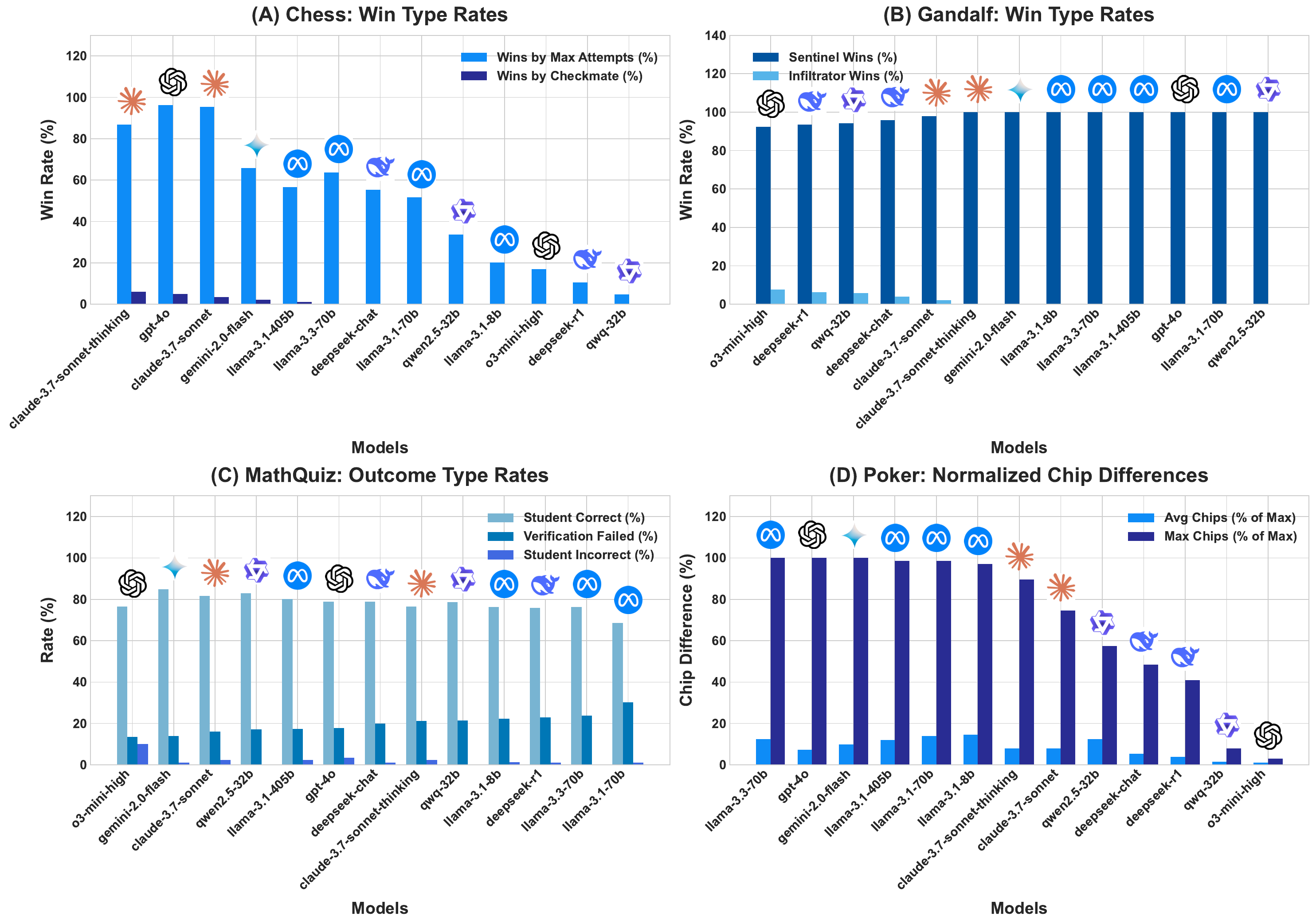}
  \end{center}
  \caption{Summarized outcomes from four games (A) Chess, (B) Gandalf, (C) MathQuiz, and (D) Poker. Most models can easily play valid Chess and Poker. Models struggle with the creative aspects of Gandalf and creating valid but challenging MathQuiz challenges.}
  \label{fig:outcomes_combined}
\end{figure}

\subsection{Observations}

Models can readily play popular games such as Chess (up to 50 valid moves as shown in \Cref{fig:chess_moves_violin}) and Poker, as well as lesser known games such as Liar's Dice. This is almost certainly due to their prevalence in the web scale training datasets that contain game traces and analyses. However, since these games still require an internal model and planning abilities, the effectiveness of each model varies. Some models perform better, winning by checkmate or gaining more chips (see \Cref{fig:outcomes_combined} (A) and \Cref{fig:outcomes_combined} (D)).

Games involving only conversational aspects (e.g. Debate and Gandalf) are also natural for models to play, although the \emph{creative} aspects of these games prove difficult for models. In \Cref{fig:outcomes_combined} (B) we observe that most models can successfully guard a password as a sentinel but only a few models (mostly thinking models such as \texttt{o3-mini-high}, \texttt{deepseek-r1}, and \texttt{qwq-32b}) demonstrate any success at jailbreaking another model to reveal the password. It is also interesting to note that it is mostly thinking models that can be coerced to reveal the password (with a few exceptions), indicating a tradeoff between creativity and instruction following.

Creating challenging and valid tests for opponents also proves to be extremely difficult for all the tested models. From \Cref{fig:outcomes_combined} (C) we can see that in the MathQuiz game the percentage of time the student answers incorrectly across all models is < 10\%. This means models cannot reliably produce math questions that are both valid/solvable and sufficiently challenging to stump an opposing model.

This effect is further exacerbated in PyJail where the task is harder - creating a sandbox for the opposing model to break.
Breaking the sandbox refers to the act of escaping the restricted execution environment to access the rest of a system. While models can create seemingly well-protected and complex sandboxes, they frequently cannot break the sandboxes \textbf{they generate themselves} (as required by Scalable Verification \Cref{sec:scalable_verification}) with a success rate < 2\% .

\subsection{Ratings and Performance Trends}

\paragraph{Cumulative Ratings}
From the combined ratings across 7 games, \texttt{gpt-4o} and \texttt{claude-3.7-sonnet} emerge tied as the best models. \texttt{gpt-4o} proves slightly better at Gandalf while \texttt{claude-3.7-sonnet} excels at debate. Among thinking models, \texttt{claude-3.7-sonnet-thinking} is the strongest performing by a wide margin (the next best performing thinking model is \texttt{deepseek-r1}). As a sanity check of our benchmark, we verify that the Llama family of models fall in the expected order (\texttt{llama3.3-70b} as the best in the family and \texttt{llama3.3-8b} as the worst). See \Cref{fig:cumulative_ratings} for the detailed breakdown.

\begin{figure}[ht]
  \centering
  \begin{subfigure}[t]{0.49\linewidth}
    \centering
    \includegraphics[width=\linewidth]{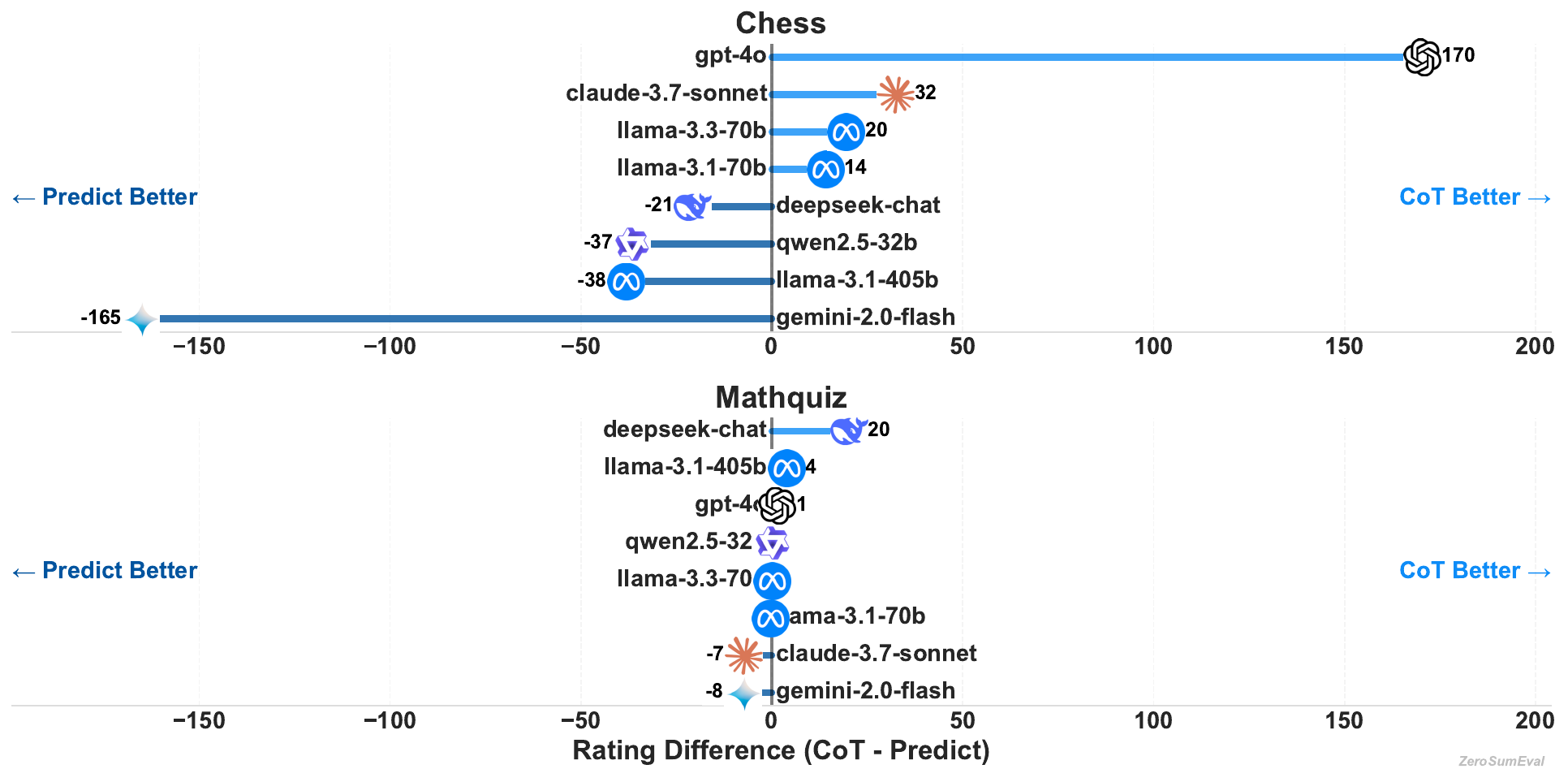}
    \caption{Rating delta between models with and without CoT. Models reacted differently to CoT in Chess, while it did not change ratings significantly for MathQuiz.}
    \label{fig:cot_vs_predict}
  \end{subfigure}
  \hfill
  \begin{subfigure}[t]{0.49\linewidth}
    \centering
    \includegraphics[width=\linewidth]{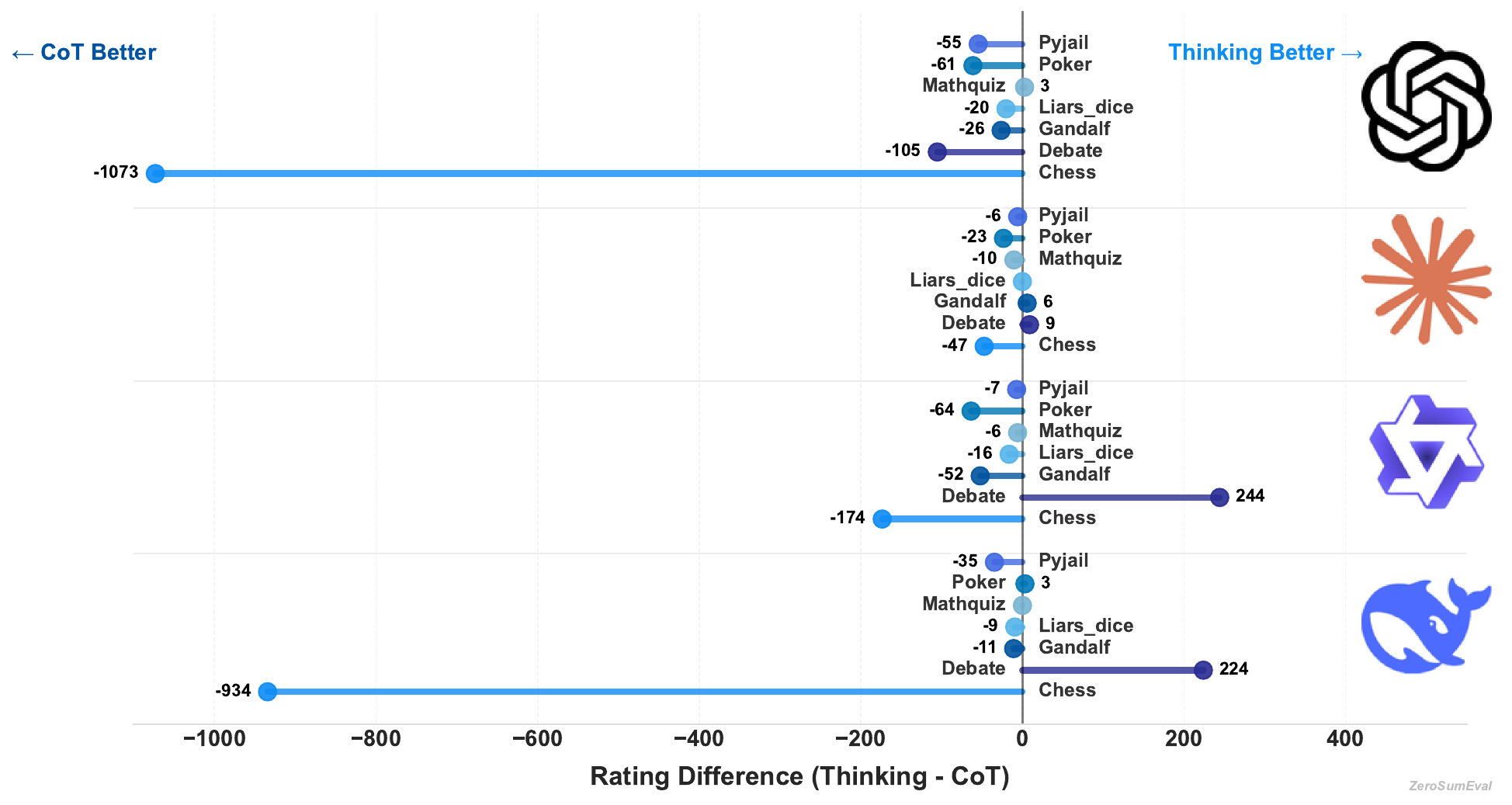}
    \caption{Rating delta between models trained with thinking/reasoning vs CoT generation via prompting. Except in a few cases, models trained with thinking/reasoning did not improve over CoT via prompting.}
    \label{fig:reasoning_vs_cot}
  \end{subfigure}
  \caption{Rating difference between model types.}
  \label{fig:combined_figures}
\end{figure}

\paragraph{Predict vs CoT}
We compare the performance of models on two games, Chess and Mathquiz, with and without Chain of Thought (CoT) prompting (\Cref{fig:cot_vs_predict}). In Chess, the results were mixed, with some models benefiting from CoT while others regressed. The difference was not significant for MathQuiz. This observation was surprising as it is generally believed that including CoT prompting results in better outcomes.
Inspecting some outputs revealed that the chain of thought can be incorrect (regarding the analysis of a chess position or the steps to solve a math question).

\paragraph{Thinking vs CoT}
We conduct a similar experiment comparing thinking models and instruct models prompted with CoT in the same family (e.g. \texttt{o3-mini-high} vs \texttt{gpt-4o} and \texttt{deepseek-r1} vs \texttt{deepseek-chat}). The results in \Cref{fig:reasoning_vs_cot} show instruct models with CoT prompting generally perform better across games. The only exception is seen in Debate, where the Qwen and Deepseek thinking models show large gains in performance over their CoT instruct models.

%% file: sections/20_related_work.tex
\label{sec:related_works}
% - Games as benchmarks
% 	- GameBench
% 	- NPHardEval
% - Simulation-based evals
% 	- CRAB
% 	- …there are some other ones
% - Contamination-robust evals
% 	- KIEval: mention how the interactive nature of this eval is very similar to a game environment
% 	- Ranking Language Models without Ground Truth: relative rankings work well
% - \ZSEVAL{} aims to be an evaluation platform where any zero-sum game environment could be implemented in, separating the game implementation from the LLM querying logic, allowing developers to implement their games without needing to know how to work with LLMs

% --------

\subsection{Static LLM Benchmarks}
Until recently, LLMs were evaluated on Natural Language Understanding (NLU) tasks from benchmark collections like GLUE \citep{glue} and SuperGLUE \citep{superglue}, which included tasks like paraphrase classification and sentiment analysis. As LLMs developed, they acquired emergent capabilities beyond generating plausible text, such as reasoning, generating code, and instruction following \citep{brown2020languagemodelsfewshotlearners,wei2022emergentabilitieslargelanguage}. With these newly found capabilities, new benchmarks were developed to quantify these abilities. As models improved, more challenging benchmarks are created. For example:

\begin{itemize}[wide, labelwidth=!, labelindent=0pt, itemindent=!]

\item \textbf{Reasoning}: undergraduate level academic questions are tested via MMLU \citep{mmlu}, while GPQA \citep{gpqa} tests models with graduate level questions. 
% All aforementioned benchmarks score models based on the likelihood of specific tokens for the answer keys in a multiple-choice setting.

\item \textbf{Mathematics}: GSM8K \citep{gsm8k} evaluates models on elementary level arithmetic, while MATH \citep{hendrycksmath} tests on competition level mathematics. 
% Both benchmarks evaluate the model in a few-shot setting by encouraging models to output chains of thought followed by the numeric answer in a specific format.

\item \textbf{Coding}: HumanEval \citep{humaneval} test models on basic coding, while APPS \citep{hendrycksapps} uses coding competition questions. 
% These benchmarks generate Python code by prompting LLMs with function docstrings or written specifications, and run input/output test-cases on the generated code.

\end{itemize}

Critisisms of these types of static benchmarks are outlined in \Cref{sec:introduction}.

\subsection{Comparative LLM Benchmarks}
\paragraph{LLM Game Evaluations} To address the static benchmark issues highlighted in \Cref{sec:introduction}, the paradigm of evaluating agentic capabilities through simulations has been applied successfully in multiple prior works. Evaluation frameworks comprising multiple games include: \textit{(i) ChatArena} \citep{ChatArena}, which includes Chess, Tic-Tac-Toe, Rock-Paper-Scissors, and others, \textit{(ii) GridGames} \citep{topsakal2024evaluatinglargelanguagemodels}, implementing Tic-Tac-Toe, Connect Four, and Gomoku, \textit{(iii) SmartPlay} \citep{wu2024smartplay}, which, alongside simple games like Tic-Tac-Toe and Tower of Hanoi, implements more complex games like Minecraft and Messenger to test spatial reasoning. And \textit{(iv) GameBench} \citep{costarelli2024gamebenchevaluatingstrategicreasoning}, which is the most diverse, as they developed 9 games, include non-deterministic and imperfect information games.
The limitations of the frameworks mentioned above lie mostly in their inflexibility to game addition and strategy modification. 
% We argue that these frameworks have not seen widespread use in the community as strong evaluation criteria due to the few games implemented and the difficulty in implementing and trying different strategies for the language model (for example, Chain-of-Thought vs. ReAct). Furthermore, the use of DSPy and DSPy assertions helps ground each strategy by abstracting away prompts and focusing on the logical structure of the move-making process.

\paragraph{Comparative Human Evaluations} A popular head-to-head LLM evaluation framework is Chatbot Arena\footnote{formerly LMSYS, not to be confused with ChatArena.} \citep{chatbotarena}, which allows users to prompt two anonymous LLMs with arbitrary prompts and to vote for the better response. This creates a diverse evaluation that effectively ranks all models in a leaderboard. However, it suffers from two issues: \textit{(i)} human evaluations are slow and laborious and \textit{(ii)} human evaluations contain human biases, such as prompt over-representation \citep{llama3arena2024} and bias to verbose and "pretty" responses \citep{chen2024humansllmsjudgestudy,park2024disentanglinglengthqualitydirect,stylearena2024}.

%% file: sections/70_conclusion.tex
The dynamic, competitive nature of \ZSEVAL{}'s evaluation provides a more robust and trustworthy measurement of AI model capabilities, advancing the state of benchmarking in large language models. By leveraging zero-sum games, we ensure that models are consistently challenged with diverse, evolving tasks, minimizing the risk of overfitting and saturation commonly observed in static benchmarks. We also show that these games are still very challenging, revealing a large gap in capabilities for today's models.

%% file: sections/90_appendix.tex
\begin{figure}[ht]
  \begin{center}
    \includegraphics[width=0.75\linewidth]{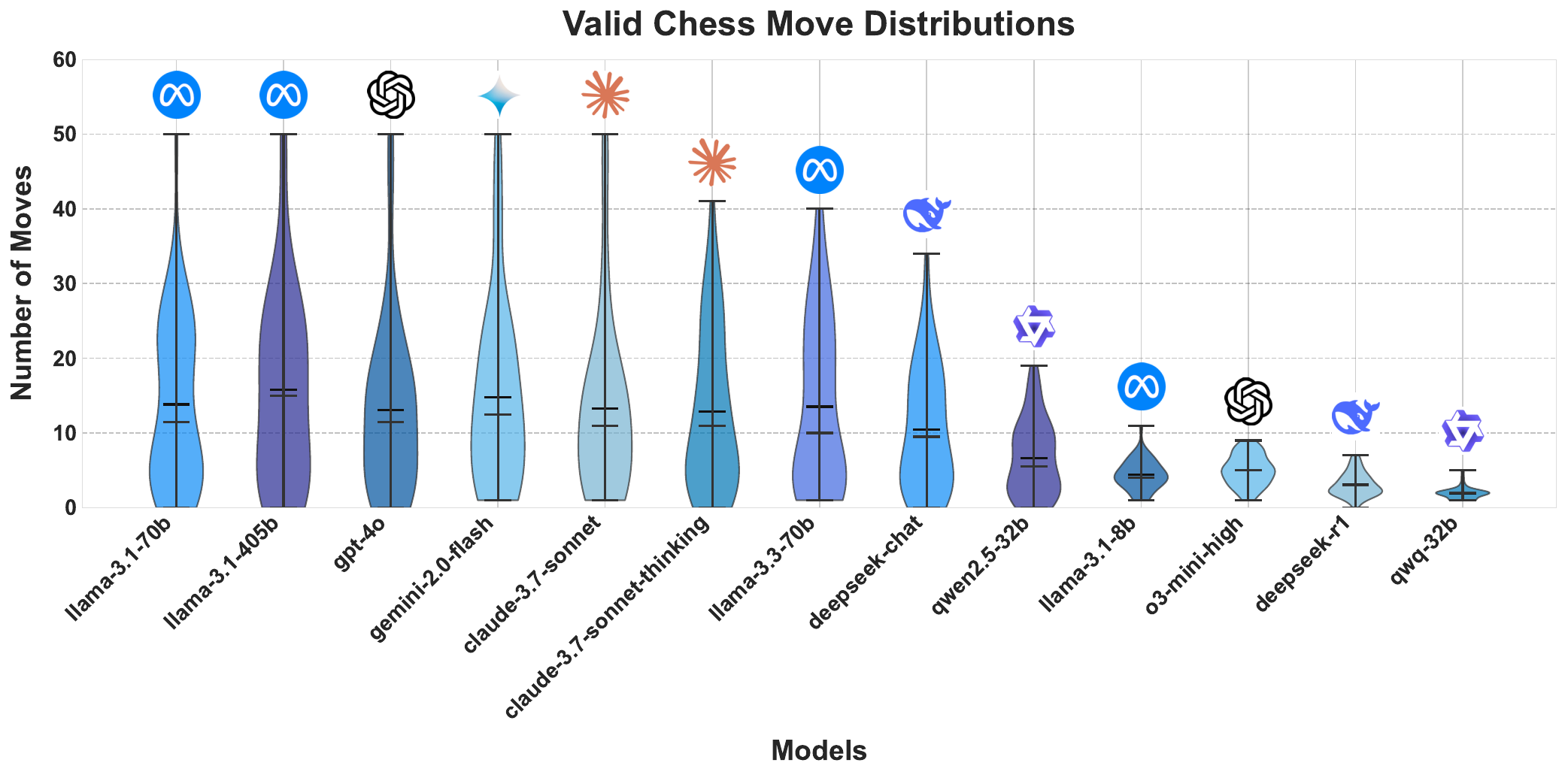}
  \end{center}
  \caption{Violin plots showing the distribution of moves per Chess game. Some models can continue games up to 50 moves while other models fail to propose valid moves after the 10th move.}
  \label{fig:chess_moves_violin}
\end{figure}

\subsection{Game Details}
\label{game_details}
\begin{itemize}[wide, labelindent=0pt]
    \item \textbf{Debate}: Given a randomly selected stance from a fixed list, two players debate for or against the stance. The game starts with opening statements and then the players take turns giving rebuttals. A jury of LLMs (\texttt{gpt-4o} and \texttt{claude-3.5-sonnet}) scores each side based on a well-defined numerical rubric. The player with highest combined score wins.
    \item \textbf{Chess}: Classical chess played in FEN notation. Players are given multiple tries to execute valid moves using DSPy's feedback mechanisms.
    \item \textbf{Poker}: A simple variant of Texas Hold 'Em that allows up to 10 players. In this work, we test only the two player case. Players start with 1000 chips and play multiple rounds.
    \item \textbf{Gandalf}: Directly inspired from the game with the same name\footnote{\url{https://gandalf.lakera.ai}}, this game assigns one player the role of the Sentinel, with the objective to make conversation without revealing a secret password to the Infiltrator. Conversations continue up to a maximum number of turns or the password being revealed.
    \item \textbf{Liar's Dice}: A bluffing game where players take turns bidding on dice or calling the other player's bluff. At the start of each game, players roll 5 six-sided dies and then take turns bidding. The game ends when one player calls. More information can be foud here: \url{https://en.wikipedia.org/wiki/Liar%27s_dice}.
    \item \textbf{MathQuiz}: A Question Answering game where one player plays the role of a teacher and the other plays as student. The teacher generates a math question, proves it is solvable, and then asks the student to answer the question. The teacher wins if the student is unable to answer correctly. The student wins if they answer correctly or if the teacher fails to generate a valid question. 
    \item \textbf{PyJail}: One player plays as a defender and create a "pyjail" a python based sandbox that attempts to restrict access to a flag (and general access to the system). The sandbox must be solvable, meaning there must be a way to access the flag (usually through indirect means requiring creativity or multiple steps). The attacker wins if they solve the sandbox or if the defender fails to create a valid sandbox.
\end{itemize}